\documentclass{article}

\usepackage{algorithm2e}

\newtheorem{Def}{Definition}

\newtheorem{Prop}{Proposition}

\usepackage{algorithmic}
\usepackage{algorithm2e}{\tiny}

\usepackage{txfonts}

\newcommand{\proof}{\emph{Proof. }}
\newcommand{\eproof}{\begin{flushright} $\square$ \end{flushright}}

\begin{document}
%

\title{Ontology alignment repair through modularization and confidence-based heuristics}
\author{Emanuel Santos, Daniel Faria, C\'atia Pesquita and Francisco Couto\\$\space$\\ 
\normalsize{LaSIGE, Faculty of Sciences of the University of Lisbon}}

\maketitle

\begin{abstract}
Ontology Matching aims to find a set of semantic correspondences, called an alignment, between related ontologies. In recent years, there has been
a growing interest in efficient and effective matching methods for large ontologies. However, most of the alignments produced for large ontologies
are logically incoherent. It was only recently that the use of repair techniques to improve the quality of ontology alignments has been explored. In this
paper we present a novel technique for detecting incoherent concepts based on ontology modularization, and a new repair algorithm that minimizes the incoherence of the resulting alignment and the number of matches removed from the input alignment. An implementation was done as part of a lightweight version of AgreementMaker system, a successful ontology matching platform, and evaluated using a set of four benchmark biomedical ontology matching tasks. Our results show that our implementation is efficient and produces better alignments with respect to their coherence and f-measure than the state of the art repairing tools. They also show that our implementation is a better alternative for producing coherent silver standard alignments.
\end{abstract}



\section{Introduction}

As ontologies became more prevalent and extensively used in domains such as biomedicine and geography, there is a growing need to automatically discover semantic correspondences between ontologies, through ontology matching \cite{David:2006:MDO:1183614.1183752, Duchateau:2009:YSM:1645953.1646311, DBLP:books/daglib/0018324, DBLP:journals/tkde/ShvaikoE13}, in order to pursue the goal of a semantic web \cite{Sabou:2008:ESW:1504492.1504503}.  This is especially relevant when a lack of coordination in ontology development results in the independent creation of ontologies for the same or related domains. The widely use Web Ontology Language (OWL) provides a way to represent ontologies with a well-defined semantics, which could include mappings between other ontologies.

In recent years, there has been a growing interest in efficient and effective matching methods for large ontologies \cite{DBLP:conf/ecai/Jimenez-RuizGZH12, Meilicke:phdthesis, gomma, DBLP:conf/ijcai/WangZX11, DBLP:conf/semweb/DuanFHKSW12}. 

The Ontology Alignment Evaluation Initiative (OAEI) \cite{DBLP:conf/semweb/EuzenatFHHMNRSSSST11} has been the major playfield for ontology alignment, with the participation of state of the art ontology matching systems in several ontology alignment challenges. After the recent introduction of the large biomedical track, an important finding of OAEI is that, although ontology matching can be seen as an offline process, some systems are not scalable enough to handle large ontologies and usually run out of memory. Another important finding is that most of the alignments produced are incoherent, i.e. lead to unsatisfable classes or properties. With respect to large ontologies alignments, the degree of incoherency is typically higher, and only one participant, LogMap \cite{DBLP:conf/semweb/Jimenez-RuizG11, DBLP:conf/ecai/Jimenez-RuizGZH12}, detects incoherencies and uses repair techniques to improve the quality of the resulting alignment. The goal of a repairing process is to restore coherency by minimally changing the input. However, reasoning-based techniques aggravate the scalability problem, which restricts their application with more effective and complex matching strategies. 

To the best of our knowledge, there are only two systems that perform alignment repair: LogMap and ALCOMO \cite{Meilicke:phdthesis}. Besides performing repair operations during the matching process, LogMap provides a repair facility that applies a local repairing process over the input alignment. This process is incomplete, i.e. it may produce an incoherent alignment, but overcomes the scalability problem. ALCOMO is a repair system that provides a complete global repair process, but fails to handle large ontologies. A more efficient incomplete process is also provided, but it continues to fail for some large ontologies alignments (see Sections \ref{sec: repair} and \ref{sec: eval} for more details).

The OAEI large biomedical track consists of finding alignments between the Foundational Model of Anatomy (FMA), SNOMED CT, and the National Cancer Institute Thesaurus (NCI). These ontologies are semantically rich and contain tens of thousands of classes. Since there is no gold standard alignment, a silver standard alignment based on the UMLS Metathesaurus \cite{ULMS} is provided for evaluating each matching problem. Repaired versions of the silver standard alignments produced by  the repair facility of LogMap and ALCOMO are also provided for evaluating the systems in competition. 

After analyzing the results provided by the repair facilities of LogMap and ALCOMO with respect to large biomedical track we identify two main problems: (1) ALCOMO and LogMap failed to repair all the incoherencies caused by disjointness restrictions between classes, which are the main cause of incoherency in alignments;  and (2) in some cases, ALCOMO and LogMap are far from minimizing the set of mappings removed from the alignments.  

In this paper, we propose a new repair algorithm that minimizes both the incoherence of the resulting alignment and the number of matches removed from the input alignment. To overcome the scalability problem, we use heuristics to determine near-optimal solutions, and filtering methods that take advantage of the confidence values of the mappings. Moreover, we introduce a modularization based technique that allows the extraction of the core fragments of the ontologies that contain only the necessary classes and relations for repairing all the incoherencies caused by disjoint restrictions.

The paper is organized as follows: Section \ref{sec:prec} describes our setting and introduces the notation used. Section \ref{sec:mod} presents our module for the extraction of core fragments and  its properties. Section \ref{sec: repair} describes our repair algorithm and main methods. Section \ref{sec: eval} presents and discusses the obtained results; and finally Section \ref{sec: conc}  is dedicated to  final remarks  and future work.

\section{Our Setting}\label{sec:prec}

We use $A,B,...X,Y,Z$ to denote classes,  $\mathcal{O}, \mathcal{O'}, \mathcal{O}_1, \mathcal{O}_2, ...$ to denote ontologies and $\mathcal{M}, \mathcal{M'}, ...$ to denote sets of mappings, also called an alignment, between classes. 

In an ontology matching setting we say that an alignment $\mathcal{M}$ between ontologies $\mathcal{O}_1$ and $\mathcal{O}_2$ is coherent if there is no class or property in $\mathcal{O}_1$ or $\mathcal{O}_2$ that is unsatisfiable due to $\mathcal{M}$ (see \cite{Meilicke:phdthesis} for a formal definition). 

With respect to superclass relations we use $A \sqsubseteq B$ and $A\sqsubseteq^d B$ to denote superclass and direct superclass relations between classes, respectively. A class $B$ is a direct superclass of $A$ if $A \sqsubseteq B$ and there is no $C$ such that $A \sqsubseteq C$, $C \sqsubseteq B$ and $B \nsqsubseteq C$. The last condition was added due to the possible existence of cycles.

We assume that ontologies are coherent and don't have cycles with respect to subclass relation between classes. The semantic inference is denoted by the symbol $\vDash$. For instance, given two ontologies $\mathcal{O}_1$ and $\mathcal{O}_2$, and a set of mapping $\mathcal{M}$, we write $\mathcal{O}_1\cup \mathcal{O}_2 \cup \mathcal{M} \vDash A \sqsubseteq B$ to denote that $A \sqsubseteq B$ is inferred with respect to the resulting merged ontology. To denote conjunction we use the symbol $\wedge$. With respect to incoherency detection, given two disjoint classes $B$ and $C$, we say that a class $A$ is incoherent if $\mathcal{O}_1\cup \mathcal{O}_2 \cup \mathcal{M} \vDash (A \sqsubseteq B) \wedge (A \sqsubseteq C)$. Since we assume that ontologies are coherent, we also say that $\mathcal{M}$ is incoherent.

Our analysis of the alignments produced for the OAEI large biomedical track by the participant ontology matching systems show that most of the incoherency found is caused by disjointness restrictions. For this reason, we only consider incoherent alignments due to subclass/disjointness conflicts. That is, when a class is subsumed by disjoint classes due to the alignment. Thus, our incoherency detection is incomplete. Moreover, as LogMap, we just consider named classes, and sub/superclass and equivalent relations between them during the incoherence detection and repair process. We followed this strategy to ensure scalability while still improving the coherency degree of the alignments.

An implementation of our algorithms was done as part of AgreementMakerLight \cite{AML}, a lightweight version of AgreementMaker \cite{AM}, a successful ontology matching platform. During the development of our algorithms we took into account the very efficient and scalable methods provided by AgreementMakerLight to explore the relationship information of the input ontologies. For instance, the cost of checking if a class is subsumed by another class becomes negligible using the AgreementMakerLight HashMaps-based data structures.

\section{Ontology Modularization} \label{sec:mod}

In order to resolve an incoherence we need to determine which mappings are culprits. The determination of all possible culprits represents a very demanding task to be performed when dealing with large ontologies. Ontology modularization techniques have been proposed and implemented to overcome the issue of scalability \cite{DBLP:conf/owled/GrauHKS07, DBLP:conf/ecai/Jimenez-RuizGZH12, DBLP:conf/cikm/DoranTI07}.

In our work we also use modularization techniques. We introduce the following extraction module that suits our repair setting.

\begin{Def}\label{def:struct}
Let $\mathcal{O}_1$ and $\mathcal{O}_2$ be ontologies, $\mathcal{M}$ a set of mappings, $\mathcal{O'}_1\subseteq\mathcal{O}_1$ and  $\mathcal{O'}_2\subseteq\mathcal{O}_2$. $\mathcal{O'}_1$ and $\mathcal{O'}_2$ are \textbf{core fragments} of $\mathcal{O}_1, \mathcal{O}_2$ and $\mathcal{M}$ if they satisfy the following conditions:

\begin{enumerate}

\item if $A$ and $B$ are disjoint classes of  $\mathcal{O}_1 \cup\mathcal{O}_2$ then $\{A,B\}\subseteq \mathcal{O'}_1 \cup \mathcal{O'}_2$;\label{cond:1}

\item if $A$ is a class and occurs in $\mathcal{M}$ then $A \in\mathcal{O'}_1 \cup \mathcal{O'}_2$;\label{cond:2}

\item if $A$ is a class of $\mathcal{O}_1 \cup \mathcal{O}_2$ such that \label{cond:3}

\begin{enumerate}

\item $\mathcal{O}_1\cup \mathcal{O}_2 \cup \mathcal{M} \vDash (A \sqsubseteq^d B) \wedge (A \sqsubseteq^d C)$, where $B$ and $C$ are distinct classes, and;\label{cond:4}

\item there is no class $D$ that satisfies $\mathcal{O}_1\cup \mathcal{O}_2 \cup \mathcal{M} \vDash (D \sqsubseteq^d B') \wedge (D \sqsubseteq^d C')$, where $B'$ and $C'$ are distinct classes, $\mathcal{O}_1\cup \mathcal{O}_2 \cup \mathcal{M} \vDash D \sqsubseteq A$ and $\mathcal{O}_1\cup \mathcal{O}_2 \cup \mathcal{M} \nvDash A \sqsubseteq D$,

\end{enumerate}

then $A\in \mathcal{O'}_1 \cup \mathcal{O'}_2$;

\item if $\mathcal{O}_1\cup \mathcal{O}_2 \cup \mathcal{M} \vDash A \sqsubseteq B$  and $\{A,B\}\subseteq \mathcal{O'}_1 \cup \mathcal{O'}_2$, then $\mathcal{O'}_1\cup \mathcal{O'}_2 \cup \mathcal{M} \vDash A \sqsubseteq B$;\label{cond:5}

\end{enumerate} 
 
We also called the \textbf{checkset} of $\mathcal{M}$ the set of classes that satisfy Condition (\ref{cond:3}).
\end{Def}

The idea behind the presented module is to compute fragments, smaller than the original ontologies, that still allow the determination of all possible culprits of incoherencies.
The module defines a set of core classes composed of: the classes that occur in a disjoint relation (Condition \ref{cond:1}) or in a mapping (Condition \ref{cond:2}), and; the classes that have more than one direct superclass and don't have a subclass with more than one direct superclass (Condition \ref{cond:3}). Condition (\ref{cond:5}) guarantees that the subclass relations between core classes are maintained. 

The following proposition shows that all the mappings responsible for incoherencies between two matched ontologies can be determined using the respective core fragments.
\begin{Prop}\label{Prop:mi}
Let $\mathcal{O}_1$ and $\mathcal{O}_2$ be ontologies, $\mathcal{M}$ a set of mappings, $\mathcal{O'}_1$ and $\mathcal{O'}_2$ the respective core fragments, $\mathcal{M'}\subseteq\mathcal{M}$, $B$ and $C$ disjoint classes. 

There is a class $A$ such that $\mathcal{O}_1\cup \mathcal{O}_2 \cup \mathcal{M'} \vDash (A \sqsubseteq B)\wedge (A \sqsubseteq C)$  if and only if there is a class $A'$ such that $\mathcal{O'}_1\cup \mathcal{O'}_2 \cup \mathcal{M'}\vDash (A' \sqsubseteq B) \wedge (A' \sqsubseteq C)$.
\end{Prop}

\proof
(``$\rightarrow$") (reductio ad absurdum) Let us assume there is $A$ such that $\mathcal{O}_1\cup \mathcal{O}_2 \cup \mathcal{M'} \vDash (A \sqsubseteq B) \wedge (A \sqsubseteq C)$ but there is no $A'$ such that  $\mathcal{O'}_1\cup \mathcal{O'}_2 \cup \mathcal{M'}\vDash (A' \sqsubseteq B) \wedge (A' \sqsubseteq C)$.  Thus, $A'\notin \mathcal{O'}_1 \cup \mathcal{O'}_2$. There are two cases:

\begin{enumerate}

\item If $\mathcal{O}_1\cup \mathcal{O}_2 \cup \mathcal{M'}\vDash B \sqsubseteq C$ (the $C \sqsubseteq B$ case is analogous) then by Conditions (\ref{cond:1}) and (\ref{cond:5}) of Definition \ref{def:struct} we have that $\mathcal{O'}_1\cup \mathcal{O'}_2 \cup \mathcal{M'}\vDash (B\sqsubseteq B) \wedge (B\sqsubseteq C)$. Contradiction.

\item Otherwise, there is a class $X$ with more than one direct superclass such that $\mathcal{O}_1\cup \mathcal{O}_2 \cup \mathcal{M'}\vDash (A \sqsubseteq X) \wedge (X\sqsubseteq B) \wedge (X\sqsubseteq C)$. If $X \in \mathcal{O'}_1 \cup \mathcal{O'}_2$ then we have a contradiction. If $X \notin \mathcal{O'}_1 \cup \mathcal{O'}_2$ then by Condition  (\ref{cond:3}) of Definition \ref{def:struct} there is a class $Y \in \mathcal{O'}_1 \cup \mathcal{O'}_2$ and $\mathcal{O}_1\cup \mathcal{O}_2 \cup \mathcal{M'}\vDash Y \sqsubseteq X$. By Condition (\ref{cond:5}) we have that $\mathcal{O'}_1\cup \mathcal{O'}_2 \cup \mathcal{M'}\vDash (Y \sqsubseteq B) \wedge (Y \sqsubseteq C)$. Contradiction. 

\end{enumerate}

(``$\leftarrow$") Trivial since $A'\in \mathcal{O}_1\cup \mathcal{O}_2$.
\eproof

Proposition \ref{Prop:mi} is mainly based on the fact that if a class is incoherent with respect to a disjoint then it must have a superclass with more than one direct superclass. Unless one of the disjoint classes subsumes the other class.

Moreover, given Proposition \ref{Prop:mi} result, a checkset (Definition \ref{def:struct}) denotes a complete set of classes to check the coherency of a mapping set wrt to disjoint restrictions. 

Table \ref{table: initial} shows the size of the core fragments computed for each of the matching problems of the OAEI large biomedical track. In all of the matching problems the size of the core fragments is significantly smaller than the original ontologies. In comparison to the module proposed by \cite{DBLP:conf/owled/GrauHKS07} and implemented by LogMap2, which computes fragments that contain 37\% of  the classes in FMA and 38\% of  the classes in NCI,  there is a considerably improvement - only 5\% of the total classes of FMA and NCI. 

Given the previous result, the checkset denotes a set of classes that need to be checked for incoherencies. This way, instead of looking of all the culprits for each incoherent class of the input ontologies, we just need to look for the culprits for each incoherent class in the checkset. Table \ref{table: initial} shows the size of the computed checkset is also significantly smaller than the size of the respective input ontologies.

\section{Alignment  Repair}\label{sec: repair}

Given an incoherent alignment, the goal of a repair procedure is to remove mappings from the input alignment in such way that the resulting set is coherent. Typically, a repair procedure ensures minimal impact on the input by, for instance, minimizing the number of removed mappings or the sum of confidence values of the removed mappings. There are two main approaches to alignment repair: global and local. 

A global repair determines the minimal impact by considering all the classes and relations of the matched ontologies. Although this approach produces better results, it is usually not scalable for large ontologies. This approach is followed by ALCOMO.

A local repair is performed by determining the minimal impact in small subsets of the matched ontologies. This approach is more efficient, but produces a bigger impact in the input alignment than the global approach. LogMap follows this approach and applies it during its ontology matching process. 

Our repairing process is divided in three main tasks: the computation of the conflict set of mappings; the filtering of conflict sets; and finally, the removal of mappings.

\subsection{Conflict sets of mappings}\label{sec:confmapp}

Our implementation takes advantage of the fragments extraction proposed in Section  \ref{sec:mod}, but also of the AgreementMakerLight data structures. 
In order to compute all the possible culprits of an incoherency, for each class in the checkset we do a full depth-first search in the core fragments structure. This way, we are able to determine all the minimal sets of mappings, called conflict sets, that are culprits of the coherencies. 

Formally, given ontologies $\mathcal{O}_1$ and $\mathcal{O}_2$, and a set of mappings $\mathcal{M}$ we compute for each checkset class $A$ and disjoint classes $B$ and $C$, the minimal set of mappings $\mathcal{M'}\subseteq \mathcal{M}$ such that $\mathcal{O}_1\cup \mathcal{O}_2 \cup \mathcal{M'}\vDash (A\sqsubseteq B) \wedge (A\sqsubseteq C)$. 

Notice that, in order to remove all the found incoherencies, we need to remove at least one mapping from each conflict set. Using a global approach, the goal is determine a minimal set of mapping that intersect all conflict sets. This way, we are able to minimize the number of removed mappings.\\

\subsection{Filtering}\label{sec:filter}

Ontology matching systems typically provide alignments with confidence values, between 0 and 1, associated to each of its mappings. These values are computed during the ontology matching and they are typically good reliability indicators. They can also be used in the repairing process when, for instance, we need to decide which mapping to remove in a conflict set.

Our repair algorithm uses that information to resolve possible ties during the selection process (see Section \ref{sec:remov}) but also uses it to perform an initial filtering of the conflict sets.
The main idea is to resolve conflict sets that appear to have a straightforward solution based on the respective confidence values. For instance, when a conflict set contains a mapping with a very low confidence value with respect to the other mappings in the set. The problem consists in establishing a value for which the lowest confidence value in a conflict set should be compared with the other confidence values. Since this value should indicate how reliable are the confidence values, we call it \textbf{confidence interval}. Thus, given a confidence interval $\epsilon$, we filter all the conflicts sets by: (1) ordering them by their highest confidence mapping, and then; (2) removing the lowest confidence mapping if there is no other mapping within its confidence interval. That is, given the lowest and the second lowest confidence values $c_1$ and $c_2$, the lowest confidence  mapping is removed if $c_1+ \epsilon <c_2- \epsilon$.

\subsection{Removing Mappings}\label{sec:remov}

Given all conflicting sets (or only part of them after filtering) we need to determine which set of mappings should be removed. The task of computing a global minimal set of mappings, which corresponds to computing a minimal set of mappings that intersect all conflicting sets, is non-scalable. For this reason we employed two main approaches: (1) compute all disjoint clusters of conflicting sets. That is, we divide the initial set of conflicting sets into sets of conflicting sets that have at least one mapping in common. 
This way we are able to determine the mappings to be removed for each of these clusters independently.  In some cases, this allows us to check if the resulting repair is in fact a global minimal. 
However, since it is not scalable approach and some matching problems have a huge number of conflict sets, it may not applicable to every case. For instance, with respect to OAEI large biomedical track and UMLS-based reference alignments (see Table \ref{table: initial}) we computed 54 and 3 initial independent clusters for the FMA-NCI and FMA-SNOMED matching problems, respectively. For the SNOMED-NCI case weren't able to employ this approach due to efficiency issues. 
(2) compute and remove the mappings that belongs to the highest number of unresolved conflict sets. This heuristics is very efficient and typically delivers the optimal solution because usually the mapping that belongs to the highest number of conflicts sets also belongs to the optimal solution. A similar strategy has been applied for repairing inconsistent databases \cite{DBLP:conf/icde/LopatenkoB07}.
However, when there are many mappings that belong to approximately the same number of conflicts sets, this heuristics fails to return the optimal solution. To overcome part of this problem, we resolve possible ties by performing a depth-first-search to determine which alternative resolves the highest number of conflict sets. The depth of this search is pre-defined. 

\subsection{The Repair Algorithm}

Algorithm \ref{alg:repair} shows a description of our repair algorithm. Its input consists of:  (1) a list of conflicting sets of mappings, $\mathcal{C}$. This list contains all the conflicting sets for a given pair of ontologies and input alignment, as described in Section \ref{sec:confmapp}. Thus, instead of taking as input the matched ontologies, the core algorithm receives the corresponding conflicting sets; (2) the initial set of mappings, $setMaps$. This set is used to keep track of the removed mappings and to be returned after the repairing process; (3) a confidence interval, $\epsilon$, for which a filtering will be performed as described in Section \ref{sec:filter}; (4) a search depth value, $sDepth$. This value establishes the depth of search when dealing with ties as described in Section \ref{sec:remov}, and finally; (5) a boolean, $disjConflicts$, that sets if the clusters of disjoint conflict sets are computed during the repair process, as described in \ref{sec:remov}.

\begin{algorithm}[h]
 \SetAlgoLined
 \textbf{Procedure:} Repair\\
\textbf{Input:} $\mathcal{C}:$ List of conflicting sets of mappings; $setMaps:$ A set of mappings; $\epsilon:$ A confidence interval;  
$sDepth:$ search depth; $disjConflicts:$ a boolean \\
\textbf{Output:} A set of repaired mappings. \\
 \begin{algorithmic} [1]
  \IF{$\epsilon\geq0$ } 
  \STATE $\mathcal{C}$ := FilterConflicts($\mathcal{C}$, $\epsilon$)
 \ENDIF
  \IF{$disjConflicts$  = true } 
  \STATE $\mathcal{P}_{\mathcal{C}}$ := DisjointConflictsSets($\mathcal{C}$)
  \ELSE
    \STATE $\mathcal{P}_{\mathcal{C}}$ := \{$\mathcal{C}$\}
 \ENDIF
 \WHILE{$|\mathcal{P}_{\mathcal{C}}|>0$}
 \STATE $\mathcal{S}$ := an element of $\mathcal{P}_{\mathcal{C}}$
  \STATE $\mathcal{P}_{\mathcal{C}} := \mathcal{P}_{\mathcal{C}}$ $\backslash$ $\mathcal{S}$
 \STATE $w$ : = WorstMapping($\mathcal{S}$, $setMaps$, $sDepth$)
  \STATE $setMaps$ : = $setMaps$ $\backslash$ $w$
  \STATE $\mathcal{S}$ : = RemoveMapping($\mathcal{S}$, $w$)
     \IF{$|\mathcal{S}|>0$ and $disjConflicts$  = true }
   	\STATE $\mathcal{P}_{\mathcal{S}}$ := DisjointConflictLists($\mathcal{S}$)
	\STATE $\mathcal{P}_{\mathcal{C}}$ := $\mathcal{P}_{\mathcal{C}} \cup \mathcal{P}_{\mathcal{S}}$
      \ELSIF{$|\mathcal{S}|>0$}
         \STATE $\mathcal{P}_{\mathcal{C}}$ := $\{\mathcal{S}\}$
   \ENDIF
 \ENDWHILE
\RETURN setMaps
\end{algorithmic}
$\space$
\caption{Description of the repair algorithm.}\label{alg:repair}
\end{algorithm}

The algorithm starts by checking if the initial filtering is performed.  If so, the method $FilterConflicts$ inputs the list of conflict sets and the confidence interval, and returns a filtered list of conflicting sets of mappings as described in Section \ref{sec:filter}.

In the case that $disjConflicts$ is set to $true$, an initial computation of the clusters of disjoint conflicting sets is performed. Notice, that this method returns a set of clusters of conflicting sets of mappings. 

Then, we enter in the main cycle of the algorithm, which will run until there is no unresolved conflicting set. In each of the steps, one cluster is selected to be resolved. In the case that $disjConflicts$ is set to $false$, $\mathcal{P}_{\mathcal{C}}$ will always contain only one element until all the conflicting sets are resolved. Given the selected cluster, the selection of which mapping to delete is performed by the method $WorstMapping$, as described in Section \ref{sec:remov}. A description of this method is shown in Algorithm \ref{alg:worst}. 

After removing the selected mapping, the conflicting sets that contain the removed mapping are marked as resolved and removed from the respective lists. This task is performed by the method $RemoveMapping$. If the $disjConflicts$ is set to $true$ a clustering process is performed over the remaining conflicting sets.\\

\begin{algorithm}
 \SetAlgoLined
 \textbf{Procedure:} WorstMapping\\
\textbf{Input:} $\mathcal{S}:$ List of conflicting set of mappings; $setMaps:$ A set of mappings; $sDepth:$ search depth\\
\textbf{Output:} The mapping to be removed. \\
 \begin{algorithmic} [1]
 \STATE $worstSet$ := $\emptyset$
  \STATE $minSim$ := $1$
    \STATE $maxCount$ := $0$
 \STATE $countMap$ := map$\langle mapping, number\rangle$ 
\FOR{ $i = 1$ to $size(\mathcal{S})$ }  
 \STATE $s$ := the $i$-$th$ element  of $\mathcal{S}$
 
 \FOR{{\bf each} $m\in s$}
  \IF{$m\in Keys(countMap)$} 
  \STATE countMap(m):= countMap(m) + 1
  \ELSE
    \STATE countMap(m):= 1
 \ENDIF
   \IF{(countMap(m) = maxCount AND Sim(m) $\leq$ minSim) OR (countMap(m) $>$ maxCount)} 
	 	\STATE $worstSet$ := \{m\}
		\STATE $minSim $ := Sim(m)
		\STATE $maxCount$ = countMap(m)
  \ELSIF{(countMap(m) = maxCount AND Sim(m) = minSim)}
  		\STATE $worstSet$ := $worstSet$ $\cup$ \{ m \}
  \ENDIF
\ENDFOR
 \ENDFOR 
  \STATE toDelete := an element of worstSet;
  \STATE maxResolved := 0
 \FOR{{\bf each} $m\in worstSet$}
   \STATE conflictsResolved = ResolvedConflicts($\mathcal{S}$, $m$, $sDepth$)
    \IF{conflictsResolved $>$ maxResolved}
    	  \STATE toDelete := m
	  \STATE maxResolved := conflictsResolved
    \ENDIF
 \ENDFOR
\RETURN toDelete
\end{algorithmic}
$\space$
\caption{Description of WorstMapping method.}\label{alg:worst}
\end{algorithm}

\section{Evaluation and Discussion}\label{sec: eval}

\begin{table*}
\centering
\caption{Number of classes in OAEI Large biomedical Track matching problems and respective core fragments and checksets.}\label{table: initial}
\begin{tabular}{|c|c|c||c|c|} \hline
 & Total & UMLS-based Align  & Core Fragments & Checkset\\ \hline
FMA - NCI & 145712 & 3024 & 7325 (5\%) & 4159 (3\%) \\ \hline
FMA - SNMD & 201452 & 9008 & 42875(21\%) & 29855 (15\%)\\ \hline
SNMD - NCI & 189188  & 18844 & 63492(34\%) & 42918 (23\%)\\ \hline
\end{tabular}
\end{table*}

In this section we identify the results produced by our implementation as AMLR.
Our evaluation was done in a server with $16Gb$ of RAM. However, all the alignments produced by AMLR can be produced using a $4Gb$ of RAM desktop without running out of memory. 

We conducted experiments using the three OAEI large biomedical track matching problems: FMA-NCI, FMA-SNMD and SNMD-NCI (see Table \ref{table: initial} for details). We also considered the UMLS-based reference alignments that are used to evaluate the OAEI competitors systems, and their repaired versions produced by ALCOMO and the repair facility of LogMap. Since the last OAEI competition, a new version of LogMap was presented, LogMap2. For this reason, we also performed the evaluation with respect to the repair facility of LogMap2. To evaluate the precision and recall more accurately, we also consider the OAEI Anatomy Track problem for which there is a more accurate and coherent reference alignment. 

With respect to the efficiency of our implementation, the time of execution is directly related to the number of conflict sets. The repair of the UMLS-based reference alignments of FMA-NCI, FMA-SNMD and SNMD-NCI, took less than 10 seconds, 15 minutes and 3 hours, respectively. The repair of the alignments produced by LogMap and LogMap2 for SNMD-NCI were executed in less than 45min. However, considering that ontology matching can be seen as an offline process, these are quite satisfactory results. 

In order to check the degree of coherency of the alignments we use the JENA API and Pellet OWL Reasoner. This is a very memory and processing intensive task, requiring the use of the 8 core 16 GB server. For instance, it took more than 10 hours on average to check the coherency of an alignment produced for the FMA - SNMD matching problem.

We divide our evaluation in two main parts: (1) we evaluate AMLR by repairing the UMLS-based alignments provided for the OAEI Large biomedical Track, and comparing the number of mappings removed and the degree of coherency with the correspondent repairs produced by LogMap, LogMap2 and ALCOMO. (2) we evaluate the precision, recall and coherency degree of AMLR by repairing the alignments produced by OMSZ for the OAEI Large biomedical and Anatomy Tracks. We also compare these results with the repairs produced by LogMap2 and ALCOMO.

\subsection{Repairing Silver Standard Alignments}\label{sec:reslsilver}

The construction of a gold or a silver standard alignment for an ontology matching problem is a very complex task. Even after several automated and manual refinements,  alignments still contain errors or incomplete information. In the case of large ontologies that problem is even bigger since manually refinement becomes impractical. The OAEI Large biomedical track uses a silver standard alignment built from the UMLS Metathesaurus. Since the resulting silver standard alignment was incoherent, repaired versions of the alignment were produced by ALCOMO and LogMap, and used to evaluate the competing matching systems. Notice that the given silver standard alignment produced does not have confidence values associated to each of the mappings. Thus, the repair algorithms can not take advantage of that information. 

In this context, we evaluate the quality of  AMLR  repairs by:  (1) determining the degree of incoherency of the alignment by counting the number of incoherent classes; (2) determining the impact in the input alignment by counting the number of removed mappings; (3) comparing its results with ALCOMO, LogMap and LogMap2; (4) using AMLR to improve the results of ALCOMO, LogMap and LogMap2.

With respect to size of the conflict sets of mappings, we computed 931, 25351 and 73515 conflict sets for FMA-NCI, FMA-SNOMED and SNOMED-NCI matching problems, respectively. Notice that, given Proposition \ref{Prop:mi} these sets include all the possible culprits of an incoherency caused by a disjoint restriction.

Table \ref{table:umlcomp} shows the result of this evaluation. 

\subsubsection{FMA-NCI}

With respect to the number of mappings AMLR and LogMap2 produce close results, with 2901 and 2902 mappings, respectively.  ALCOMO removes 80 mappings more. However, with respect to incoherency, ALCOMO produces a repair with only 10 incoherent classes, the same number as AMLR. LogMap and LogMap2 produce alignments with a high number of incoherent classes. Thus, AMLR produces the best results with respect to number of mappings removed and the coherence degree. 

To show that the repaired alignment provided by the other systems could be improved by AMLR, we also repaired their respective alignments. The results show that AMLR considerably improves the incoherence degree of LogMap and LogMap2 by reducing it to 10 incoherent classes, as AMLR and ALCOMO. Moreover, AMLR produce optimal and near-optimal repairs for LogMap and LogMap2 repaired alignments, respectively. This was possible by applying the cluster strategy described in Section \ref{sec:remov}.

With respect to ALCOMO, AMLR did not remove any mappings, which was expected since ALCOMO already had the same number of incoherent classes as AMLR. 

Moreover, we were able to produce an optimal repair for LogMap case, and, at least, near-optimal minimal repairs  for the remaining alignments produced by LogMap, LogMap2 and ALCOMO.

\subsubsection{FMA-SNMD}

With respect to the number of mappings AMLR produces by far the best results, with 8349 mappings. The second best is ALCOMO with 8132. With respect to incoherency, AMLR is the only one that produces a fully coherent alignment. Moreover, only ALCOMO produces a comparable lower number of incoherent classes. LogMap and LogMap2 did not produce a quality alignment. 

In this case we also repaired the resulting alignments of the other systems. In all of the cases we are able to considerably improve their results.  For instance, by removing 6, 4 and 14 mappings from the LogMap, Logmap2 and ALCOMO alignments, respectively, we were able to achieve fully coherent alignments.

\subsubsection{SNMD-NCI}\label{sec: snmd-ncicase}

The SNMD-NCI task is very demanding in terms of memory, so both ALCOMO and our incoherency check were unable to provide results. This was excepted since the UMLS-based alignment for this matching problem has more than double the number of mappings with respect to the FMA-SNMD case, which already took an average of 10 hours to verify the coherency of each alignment.

Nevertheless, with respect to the number of mappings AMLR produced an alignment with less mappings than LogMap and LogMap2. However, by applying AMLR over the repairs produced by LogMap and LogMap2 we also obtained a lower number of mappings. Given the results of FMA-NCI and FMA-SNMD cases, this  indicates that those alignment have a much higher degree of incoherence. For instance, AMLR removes 324 mappings from the LogMap alignment, which indicates that the majority of the incoherencies found by AMLR were still in LogMap alignment.

\begin{table*}
\centering
\begin{tabular}{|c|c|c||c|c||c|c|} \hline
 & \multicolumn{2}{|c||}{FMA - NCI}  & \multicolumn{2}{c||}{FMA - SNMD}   &  \multicolumn{2}{c|}{SNMD - NCI}\\ \cline{2-7}
 & Nm & Inc & Nm & Inc & Nm & Inc \\ \hline
AMLR & 2901 & 10 & 8349 & 0 & 18065 &- \\ \hline
LogMap1 & 2898 & 7867& 8111& 61334 & 18324&- \\ \hline
LogMap1+AMLR& 2882 & 10  &  8095 & 0 & 18000 &- \\ \hline
LogMap2 & 2902  & 16399  &8096 & 27250 & 18128 &-\\ \hline
LogMap2+AMLR& 2877 & 10 & 8092& 0 & 17796&- \\ \hline
ALCOMO & 2819 & 10 & 8132 & 92 & NA &-\\ \hline
ALCOMO+AMLR& 2819 & 10 & 8118& 0 & NA&- \\ \hline
\end{tabular}
\caption{Evaluation of repairs produced for the UMLS-based reference alignments used in OAEI Large biomedical Track. \emph{Nm} denotes the number of mappings and \emph{Inc} denotes the number of incoherent classes. X+AMLR represents the results of applying AMLR over the results of X.}\label{table:umlcomp}
\end{table*}

$\space\\$

This evaluation clearly shows that AMLR obtains the best results with respect to the impact on the input alignment, and with respect to incoherency.  Moreover, they also show that AMLR provides a better alternative for obtaining a more accurate silver standard alignment for the OAEI Large biomedical Track. 

\subsection{Repairing alignments}

Besides ensuring the coherency of an alignment, a repair procedure is also used to improve the quality of alignment in terms of f-measure. Since by its nature the repair procedure can not improve the recall, its goal is to improve precision without decreasing recall. Thus, its application produces better results when the input alignment has low precision.

To evaluate the impact of AMLR on the f-measure of the input alignments we consider the alignments produced by OMSZ for OAEI Anatomy  and Large biomedical Tracks. With respect to the Anatomy track we use the gold standard alignment provided, which is coherent and regarded as accurate. In this case we use an alignment produced by OMSZ in an initial phase of its matching process, where precision is low. With respect to the Large biomedical track, since we show in Section \ref{sec:reslsilver} that AMLR produces much better results than the remaining systems, we used the repaired alignments produced by AMLR for the UMLS-based alignments as the reference alignments. 

We also evaluate the results of an initial filtering of the conflicting sets as described in Section \ref{sec:filter}. For this purpose, we compare the results of AMLR with four different settings: no filtering, and filtering with confidence intervals of 0.1, 0.05 and 0.0, respectively. Notice that with the confidence interval set to 0.0 all the conflicting sets that have mappings with distinct confidence values will be filtered. 

Tables \ref{table:psm}, \ref{table:fma-nci-small} and \ref{table:snomed-fma-small} show the results of this evaluation.

\subsubsection{Anatomy}

Given the coherency degree of the resulting alignments (see Table \ref{table:psm}), we conclude once again that AMLR produces the best results, with 0 incoherent classes. LogMap2 produces an alignment with almost as many incoherent classes as the initial non-repaired alignment.  

With respect to f-measure values, since the initial alignment is small the resulting values are closer to the initial alignment values. However, it is AMLR who produces the best results, 67.1\% f-measure in one of its settings, which represents a significant  0.7\% improvement over the initial f-measure value. Notice that the worst of the four settings of AMLR has the same f-measure as ALCOMO, 66.7\%, and still a better f-measure than LogMap2, 66.6\%.

\begin{table}[h]
\centering
\begin{tabular}{|c|c|c|c|c|} \hline
 & Precision & Recall  & F-measure & Coherence\\ \hline
Not repaired & 59.3 & 74.3 & 66.4 & 5006 \\ \hline
AMLR (no filter) & 60.0 & 74.2 & 66.7 & 0 \\ \hline
AMLR (0.1) & 60.3 & 74.3 & 66.9 & 0 \\ \hline
AMLR (0.05) & 60.6 & 74.2 & 67.0 & 0 \\ \hline
AMLR (0.0) & 60.7 & 74.1 & 67.1 & 0 \\ \hline
LogMap2 & 59.6 & 74.3 & 66.6 & 4998 \\ \hline
ALCOMO & 59.9 & 74.2 & 66.7 & 2 \\ \hline
\end{tabular}
\caption{Evaluation of the repairs produced for an initial-phase alignment of OMSZ wrt OAEI Anatomy Track.}\label{table:psm}
\end{table}

\subsubsection{FMA-NCI}

In this evaluation (see Table \ref{table:fma-nci-small}) ALCOMO and AMLR produced similar results with respect to coherency and f-measure. These results were excepted since the initial alignment already had a high precision value. Both ALCOMO and AMLR produce alignments with 83.8\% f-measure or more, and with only 2 incoherent classes. LogMap2 produces the worst results by producing an alignment with 147 incoherent classes and the lowest f-measure.

\begin{table}[h]
\centering
\begin{tabular}{|c|c|c|c|c|} \hline
 & Precision & Recall  & F-measure & Coherence\\ \hline
Not repaired & 96.6 & 72.4 & 83.4 & 248 \\ \hline
AMLR (no filter) & 97.5 & 72.1& 83.8 & 2 \\ \hline
AMLR (0.1) & 97.5 & 72.1 & 83.8 & 2 \\ \hline
AMLR (0.05) & 97.8 & 72.2 & 84.0 & 2 \\ \hline
AMLR (0.0) & 97.5 & 71.5 & 83.5 & 2 \\ \hline
LogMap2 & 97.0 & 71.8 & 83.4 & 147 \\ \hline
ALCOMO & 97.4 & 72 & 83.8 & 2 \\ \hline
\end{tabular}
\caption{Evaluation of the repairs produced for the alignment of OMSZ wrt OAEI FMA-NCI matching problem.}\label{table:fma-nci-small}
\end{table}

\subsubsection{FMA-SNOMED}

In this case ALCOMO didn't finish after 10 hours and, thus, didn't provide any result. The results show (see Table \ref{table:snomed-fma-small}) that the different settings of OMSZ may produce very distinct results. For instance, by not applying any filter, OMSZ produces an alignment with a f-measure 1.1\% higher than the initial alignment. However, by applying a filter with a confidence interval of 0.0 or 0.05, OMSZ produces a worst alignment with respect to f-measure. LogMap2 also produces an alignment with a lower f-measure than the initial alignment.

The contrasting results produce by the different confidence intervals can be explained by the number of conflicting sets filtered. In this case, we filter 2845, 6398, 13832 conflicting sets on a total of 13932 with respect the confidence intervals of $0.1$, $0.05$ and $0.0$, respectively. In the case of a confidence interval of 0.0 most of the conflicting sets were filtered. Thus, given the high number of conflicting sets, this filtering produced an alignment with a lower recall.  

With respect to the coherency degree, as in the previous cases, OMSZ produces much better results than the initial alignment and LogMap2.

\begin{table}[h]
\centering
\begin{tabular}{|c|c|c|c|c|} \hline
 & Precision & Recall  & F-measure & Coherence\\ \hline
Not repaired & 89.6 & 68.3 & 78.3 & 12369 \\ \hline
AMLR (no filter) & 93.5 & 67.5 & 79.4 & 18 \\ \hline
AMLR (0.1) & 93.8 & 66.6 & 79.0 & 8 \\ \hline
AMLR (0.05) & 94.0 & 63.1 & 77.0 & 36 \\ \hline
AMLR (0.0) & 94.6 & 51.7 & 69.9 & 19 \\ \hline
LogMap2 & 90.3 & 66.4 & 77.4 & 242 \\ \hline
ALCOMO & NA & NA & NA & NA \\ \hline
\end{tabular}
\caption{Evaluation of the repairs produced for the alignment of OMSZ wrt OAEI FMA-SNOMED matching problem.}\label{table:snomed-fma-small}
\end{table}

\subsubsection{SNOMED-NCI}

In this case ALCOMO ran out of memory and, thus, didn't provide any result. As in Section \ref{sec: snmd-ncicase} we were not able to determine the coherency degree of the alignments. 

The results show (see Table \ref{table:snomed-nci-small}) that OMSZ produces better results than the initial alignment. Its best settings produce an alignment with a 0.9\% f-measure improvement. LogMap2 also improves the f-measure of the initial alignment, but by just 0.3\%.

\begin{table}[h]
\centering
\begin{tabular}{|c|c|c|c|c|} \hline
 & Precision & Recall  & F-measure & Coherence\\ \hline
Not repaired & 89.0 & 61.2 & 73.8 & - \\ \hline
AMLR (no filter) & 91.6 & 60.9& 74.7 & - \\ \hline
AMLR (0.1) & 94.2 & 59.2 & 74.7 & - \\ \hline
AMLR (0.05) & 94.6 & 57.4 & 73.8 & - \\ \hline
AMLR (0.0) & 91.6 & 60.9 & 74.7 & - \\ \hline
LogMap2 & 90.3  & 60.8 & 74.1 & - \\ \hline
ALCOMO & NA & NA & NA & - \\ \hline
\end{tabular}
\caption{Evaluation of the repairs produced for the alignment of OMSZ wrt OAEI SNOMED-NCI matching problem.}\label{table:snomed-nci-small}
\end{table}

$\space\\$

With respect to different settings of AMLR tested, the results show that is not clear how to set the confidence interval. However, it is clear that by filtering the conflicting sets we can obtain better and more efficient results.

\section{Conclusions and Future Work}\label{sec: conc}

In this paper we presented a new modularization based technique to extract the core fragments of the ontologies  involved in alignment incoherencies, and a new repair algorithm that uses heuristics and filtering strategies to determine near-optimal solutions  to provide a coherent alignment.

We did an extensive evaluation where we compared our implementation to the state of the art repairing systems. The results show that our repair implementation produces better results with respect to coherency, i.e. number of incoherent classes, and impact in the input alignment, i.e the number of mappings removed.
In fact, our implementation produced remarkably better results than the repaired silver standard alignments of the OAEI Large biomedical Track. Thus, proving to be a better alternative for producing coherent silver standard alignments. 

The results also show that our filtering strategy can obtain good results when mappings are associated with confidence values. However, the selection of an optimal confidence intervalf is not straightforward. 

As ongoing work, we are adding parallel strategies to our implementation to take advantage of the current multi-core computers, and, hence, to improve the efficiency of the repairing process. The increase in the efficiency could also be used to achieve better results by, for instance, performing a deeper search when looking for the mapping to be removed. We are also integrating our repair algorithm in OMSZ. Our aim is to create a repair module in OMSZ that can be called during the matching process, to overcome the loss of recall caused by applying repair on the final alignment only.

As for future work, we want to consider for repair other restrictions and properties between classes besides disjoint restriction (e.g. allValuesFrom and someValuesFrom OWL restrictions).


%
\bibliographystyle{abbrv}
\bibliography{mypaper}  

\end{document}